\documentclass[10pt,twocolumn,letterpaper]{article}

\usepackage[algorithms]{wacv}      

\usepackage{graphicx}
\usepackage{amsmath}
\usepackage{amssymb}
\usepackage{booktabs}
\usepackage{multirow}
\usepackage{colortbl} 
\usepackage{xcolor}   
\usepackage{tabularx} 
\definecolor{lightgray}{gray}{0.8} 

\usepackage[pagebackref,breaklinks,colorlinks]{hyperref}

\usepackage[capitalize]{cleveref}
\crefname{section}{Sec.}{Secs.}
\Crefname{section}{Section}{Sections}
\Crefname{table}{Table}{Tables}
\crefname{table}{Tab.}{Tabs.}

\def\wacvPaperID{1588} 
\def\confName{WACV}
\def\confYear{2025}

\begin{document}

\title{CoAPT: Context Attribute words for Prompt Tuning}

\author{
    \makebox[\textwidth][c]{Gun Lee$^1$ \hspace{20pt} Subin An$^1$ \hspace{20pt} Sungyong Baik$^2$ \hspace{20pt} Soochahn Lee$^1$} \\
    \makebox[\textwidth][c]{Kookmin University$^1$, Hanyang University$^2$} \\
    \makebox[\textwidth][c]{\tt\small \{leegun4488, nesquiq, sclee\}@kookmin.ac.kr, dsybaik@hanyang.ac.kr}
}
\maketitle
\begin{abstract}
\textcolor{black}{
We propose a novel prompt tuning method called \emph{CoAPT (Context Attribute words in Prompt Tuning)} for few/zero-shot image classification. 
The core motivation is that attributes are descriptive words with rich information about a given concept.
Thus, we aim to enrich text queries of existing prompt tuning methods, improving alignment between text and image embeddings in CLIP embedding space.
To do so, \emph{CoAPT} integrates attribute words as additional prompts within learnable prompt tuning and can be easily incorporated into various existing prompt tuning methods.
To facilitate the incorporation of attributes into text embeddings and the alignment with image embeddings, soft prompts are trained together with an additional meta-network that generates input-image-wise feature biases from the concatenated feature encodings of the image-text combined queries.
Our experiments demonstrate that \emph{CoAPT} leads to considerable improvements for existing baseline methods on several few/zero-shot image classification tasks, including base-to-novel generalization, cross-dataset transfer, and domain generalization. 
Our findings highlight the importance of combining hard and soft prompts and pave the way for future research on the interplay between text and image latent spaces in pre-trained models.
}
\end{abstract}

\thispagestyle{plain} 
\pagestyle{plain} 
\section{Introduction}
\label{sec:intro}


Vision-language pre-trained models have fueled remarkable progress in few/zero-shot image classification~\cite{CLIP,ALIGN,CoOp,CoCoOp}. 
Models such as CLIP\cite{CLIP} or ALIGN\cite{ALIGN} are trained by contrastive learning of interrelations between pairs of text and image, and demonstrate generalization to various downstream tasks.
For image classification as a downstream task, class-specific text embeddings given by the text encoder from manually crafted prefixes, such as ``a photo of a \{class name\}.'' are used to generate queries that are subsequently matched with the image embeddings processed through the image encoder~\cite{CoOp,CoCoOp}. 

Transfer learning by fine-tuning is perhaps the most direct approach to adapt the pre-trained model to the particular downstream task~\cite{CLIP}. 
But this can be computationally intensive and risks degrading pre-trained knowledge of the model~\cite{zheng2023preventing}.
Recent works~\cite{CoOp, CoCoOp} have introduced supplementary prompts to existing vision language models (VLM), with a fixed backbone encoder.
The additional prompts enrich the context to enhance the correlation of the embedded tokens of corresponding text and image pairs.

In few-shot classification, these prompts can be categorized into two types: hard and soft. 
Hard prompts are text tokens encoded from manually crafted word sets~\cite{shin2020autoprompt}. 
Soft prompts are token vectors in the high-dimensional, continuous latent space, optimized through a learning process~\cite{CoOp}.
By expanding the model's dimensions, soft prompts may offer a more flexible and potentially more nuanced way for the model to adapt to the given task without the need for labor-intensive manual text crafting.


The positive impact of soft prompts depends on the design of the prompt and its optimization process.
This has led to many works~\cite{CoOp,CoCoOp,PromptSRC} exploring various scenarios for initialization, optimization, and configuration of soft prompts for few/zero-shot classification.
\textcolor{black}{
However, such methods increase the training difficulties and hinder the model explainability.
Thus, we turn our attention back to designing hard prompts that can better facilitate soft prompt learning, without a significant increase in human effort or compromising explainability. 
}

\begin{figure*}[tb]
  \centering
  \includegraphics[width=\textwidth]{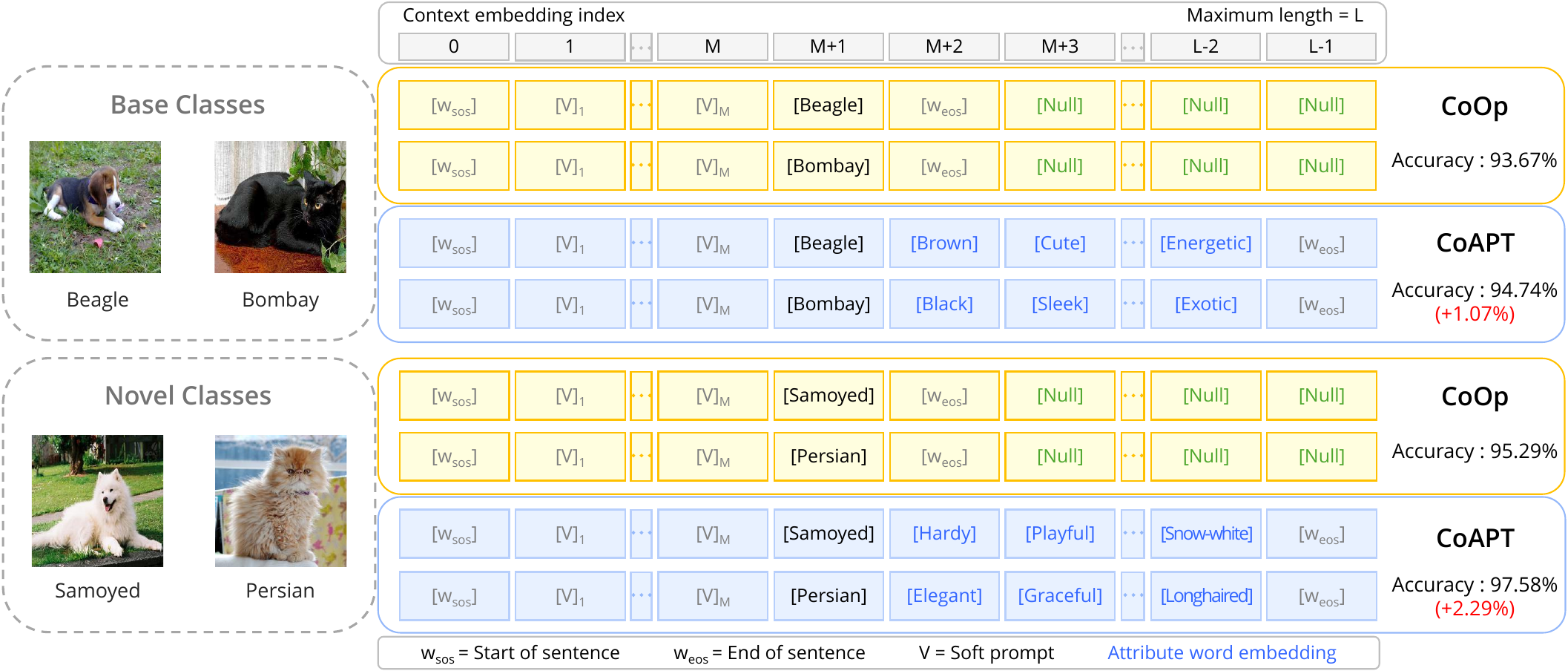}
  \caption{\textbf{Comparative overview of CoAPT.} Existing soft prompt tuning methods, such as CoOp~\cite{CoOp}, do not fully utilize the text encoder input. The empty slots in the text query can be enhanced by integrating additional hard prompts. For the sample classes from the OxfordPets dataset\cite{oxfordpets}, CoAPT achieves better classification accuracy for both base and new classes.}
  \label{fig_overview}
\end{figure*}

In this paper, we propose a novel method, termed CoAPT (Context Attribute words in Prompt Tuning), for incorporating hard prompts with soft prompt tuning. 
The core of CoAPT is the addition of {\emph{attribute words}} as hard prompts in tuning soft prompts, as depicted in Figure~\ref{fig_overview}. 
We believe that our work provides important contributions:
\begin{itemize}
    \item CoAPT is simple, easy to implement, and can be integrated with many existing prompt tuning methods.
    \item CoAPT demonstrates consistent empirical improvement upon baselines across base-to-novel generalization, cross-dataset evaluation and domain generalization tasks.
    \item We present comprehensive ablative evaluations and analysis of different variations of CoAPT, with emphasis on the process and configurations for generating relevant hard prompts.
\end{itemize}

\section{Related Work}

\noindent
\textbf{Vision-Language Models (VLMs).}
VLMs, including CLIP~\cite{CLIP}, ALIGN~\cite{ALIGN}, LiT~\cite{Lit}, Filip~\cite{yao2021filip}, and Florence~\cite{Florence}, have demonstrated capabilities for joint learning of images and natural language to generate multi-modal latent representation embeddings.
The core technology behind VLMs includes large-scale training data and contrastive learning methods.
Dual encoders for vision~\cite{vit,resnet} and text~\cite{transformer,brown2020language,kenton2019bert} are jointly trained on large scale datasets of paired images and text descriptions.
By aligning the embedding vectors of image and text from the respective encoders during training, VLMs learn the interrelations between two different modalities.
As the number of aligned pairs increase with larger datasets, the learned interrelations become more sophisticated, enabling more relevant cross-modal generations, as demonstrated by GPT-4-vision~\cite{achiam2023gpt} and Gemini-Pro~\cite{team2023gemini}.

\noindent
\textbf{Prompt Learning in Natural Language Processing}
originated from the idea of appending prefixes as language instructions to the input text.
This method enabled pre-trained LLMs like GPT-3,4\cite{brown2020language,achiam2023gpt}, Llama\cite{Llama} to adapt to downstream tasks in few-shot or zero-shot settings.
Subsequent research focused on constructing more effective prompts. 
Researchers have explored generating optimal prompts through methods like text mining\cite{jiang2020can}, paraphrasing\cite{jiang2020can,haviv2021bertese,yuan2021bartscore}, and gradient-based approaches\cite{li2021prefix,lester2021power,shin2020autoprompt,zhong2021factual,qin2021learning,hambardzumyan2021warp}. 
These advances demonstrate the potential of prompt learning for robust transfer learning.

\noindent
\textbf{Prompt Learning in Vision-Language Model.}
In CLIP~\cite{CLIP}, pre-defined prefixes are used as hard prompts to evaluate the pre-trained model on several zero-shot tasks. 
Subsequent methods~\cite{CLIP,shu2022test,gao2021making} employed a global soft prompt together with the input for the target tasks.
Further methods were proposed using sets of soft prompts, trained per input class or task~\cite{CoOp,ju2022prompting,shen2024multitask}, or trained to generate input-dependent prompts~\cite{CoCoOp,jung2023generating}.
These methods can also be categorized by whether the prompts are for text or image features.
In CoOp~\cite{CoOp} and CoCoOp~\cite{CoCoOp}, the soft prompts are defined for the text encoder.
In MaPLe~\cite{MaPLe}, VPT~\cite{VPT}, and PromptSRC~\cite{PromptSRC}, trainable soft prompts are defined for the image encoder.
%

\noindent
\textbf{Augmenting Prompt Tuning in Few/Zero-Shot Classification.}
The research community has explored various methods to generate task-specific text prompts, enhancing model adaptability in few/zero-shot scenarios. 
In~\cite{zheng2023exif,parashar2023prompting}, they utilize pre-existing metadata like EXIF of images or scientific name of class as their text prompt.
Studies such as~\cite{ArGue,cupl,kaul2023multi,liu2023democratizing,esfandiarpoor2023follow,maniparambil2023enhancing} explored leveraging the inherent knowledge within foundation models.
Their methods incorporated automated prompt generation per task, class, or instance, with pre-trained foundation models.
In~\cite{cupl}, LLMs are instructed to generate descriptions of the class object, which are directly used as prompts without soft prompts.
In~\cite{ArGue}, LLM responses go through additional prompt mining processes, including manual curation, for better alignment with input image features.
In~\cite{an2024perceptionclip}, the use of CLIP inferred per-image attributes such as background or orientation as query context is explored.


\begin{figure*}[!]
  \centering
  \includegraphics[width=\textwidth]{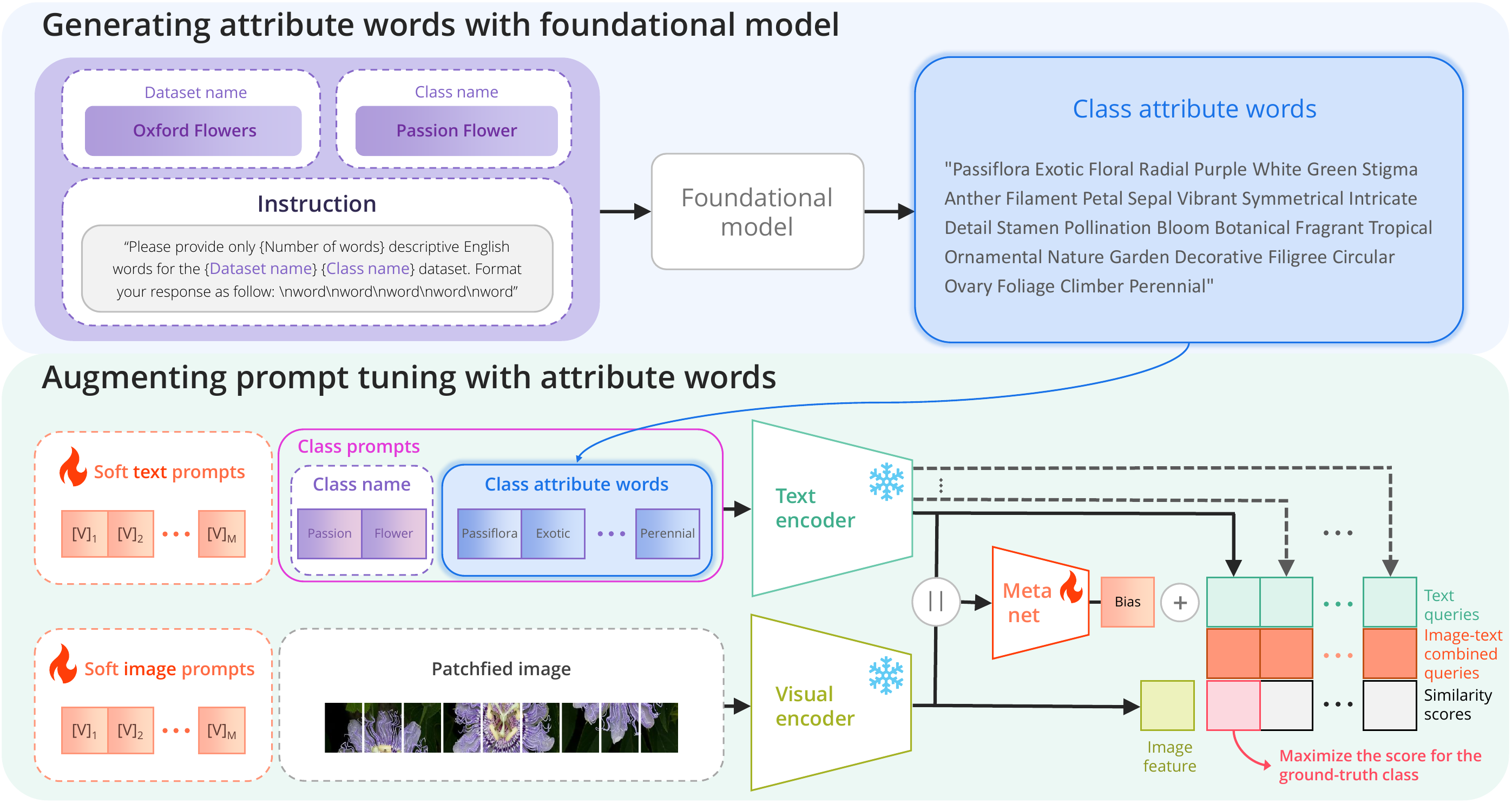}
  \caption{\textbf{Overview of CoAPT method with baseline prompt learning model.} The CoAPT method consists of two steps. First, attribute words are generated using a language model, which is a one-time process. Second, during prompt learning, these words are combined with the soft prompt and class token. Inputs generate queries processed by a Meta-network, adding a bias term to the text queries. The combined image-text queries are then used to maximize the score for the ground-truth class.}
  \label{coapt}
\end{figure*}

\section{Method}
\subsection{Preliminaries}

\noindent
\textbf{Prompt Engineering.}
CLIP~\cite{CLIP} has provided a text and image encoder pair that is highly correlated with an open vocabulary.
Naturally, CLIP has been successfully applied for image classification.
Given an image dataset comprising $|C|$ classes and a specified class name word embedding denoted as
$\mathbf{c}_{i}$, $i = \{1,...,|C|\}$, a text prompt $\mathbf{y}_{i} = \{\mathbf{w}_{SOS}, \mathbf{w}_{1}, \mathbf{w}_{2}, \cdots, \mathbf{c}_{i}, \mathbf{w}_{EOS}\}$ is crafted as the embedding of the context phrase with the class name, such as ``a photo of a \{class name\}''. Here, $\mathbf{w}_{j}$ are the embeddings of the phrase words, with $\mathbf{w}_{SOS}$ and $\mathbf{w}_{EOS}$ denoting the start and end of the sentence respectively.

This prompt is then processed by the CLIP text encoder, denoted as $g_T$, to produce a class-specific text feature encoding $\mathbf{t}_{i} = g_T(\mathbf{y}_{i})$. 
Simultaneously, an image $\mathbf{x}$ to be classified is passed through the model's image encoder, $g_V$, to generate an image-specific feature, $\mathbf{f} = g_V(\mathbf{x})$. 
The classification of the image into one of the $|C|$ categories is then determined by computing a probability distribution:
\begin{equation}
P(y=i|\mathbf{x}) = \frac{\exp \left( \cos \left( \mathbf{f}, \mathbf{t}_i \right) / \tau \right)}{\sum_{j=1}^{|C|} \exp \left( \cos \left( \mathbf{f}, \mathbf{t}_j \right) / \tau \right)},
\label{eq_prob_dist}
\end{equation}
over the class labels $y = \{1,2,\cdots,|C|\}$, based on the similarities between per-class $\mathbf{t}^{i}$ and $\mathbf{f}$.
The parameter $\tau$ represents the temperature in the CLIP model, controlling the range of logits in the softmax function.

\noindent
\textbf{Soft Prompt Learning.}
Adaptation of CLIP models can be challenging because the original information can be lost as training progresses. Additionally, the model's large number of parameters can lead to overfitting, especially when using a small dataset.
Using prompt learning, the model can be readily adapted to various settings with only a small portion of additional parameters, while preserving pre-trained weights.

Soft prompts are trainable continuous tokens, which can be defined for both the text encoder and image (vision) encoder, denoted here as $\mathbf{P}^T = \{\mathbf{p}^T_m\}_{m=1}^M$, $\mathbf{P}^V = \{\mathbf{p}^V_m\}_{m=1}^M$, respectively.
By learning the best soft prompts for the given dataset, they can be effective for classification by complementing the handcrafted prompts $\mathbf{y}_{i}$. 
They can be easily concatenated for both text as $\tilde{\mathbf{y}}_{i} = \{\mathbf{w}_{SOS}, \mathbf{p}^{T}_{1}, \cdots, \mathbf{p}^{T}_{M}, \mathbf{c}_{i}, \mathbf{w}_{EOS}\}$, and image as $\tilde{\mathbf{x}} = \{\mathbf{x},\mathbf{p}^{V}_{1}, \mathbf{p}^{V}_{2}, \cdots, \mathbf{p}^{V}_{M}\}$. 
This results in augmented feature embeddings $\tilde{\mathbf{t}}_i = g_T(\tilde{\mathbf{y}}_{i})$ and $\tilde{\mathbf{f}} = g_V(\tilde{\mathbf{x}})$.
The probability distribution over the class labels $l_{i}$ is also computed by Eq.(\ref{eq_prob_dist}), but with input $\tilde{\mathbf{t}}_i$ and $\tilde{\mathbf{f}}$ in place of ${\mathbf{t}}_i$ and ${\mathbf{f}}$.

\subsection{Context Attribute Words for Prompt Tuning}

\noindent
\textbf{Context Attribute Words.}
Whether the contexts are phrase words or soft prompts, they are invariably less than the total permitted context length in many existing methods~\cite{CoOp,CoCoOp,MaPLe,PromptSRC,ProMetaR,DePT}, resulting in suboptimal prompts and thus failing to fully exploit the capability of VLMs, as described in Figure~\ref{fig_overview}.
To provide rich relevant information instead of empty slots and thereby take full advantage of prompt and VLM capability, we fill these empty context slots with words that are highly correlated with a given class: namely, attributes of the class.
While there may be many different ways to collect these words, in this work, we apply a large language model (LLM) to generate these words to minimize manual human efforts.
We design a template instruction phrase that is customized with the dataset and class names, on which LLM is conditioned to generate the attribute words $\mathbf{A}_i = \{\mathbf{a}^n_i\}^N_{n=1}$.
As attribute words are generated for each class, they are denoted as $\mathbf{a}^{n}_{i}$.
Then, the $i_{th}$ class text prompt is augmented with $\mathbf{A}_i$, as $\tilde{\mathbf{y}}^{a}_{i} = \{\mathbf{w}_{SOS}, \mathbf{p}^{T}_{1}, \cdots, \mathbf{p}^{T}_{M}, \mathbf{c}_{i}, \mathbf{a}^{1}_{i}, \cdots, \mathbf{a}^{N}_{i}, \mathbf{w}_{EOS}\}$.
The attribute words provide enhanced context, improving discriminative representation from soft prompt tuning. 
We show sample attribute words in the supplemental material.

\noindent
\textbf{Input Image Specific Query Adaptation.}
Depending on the resulting tuned soft prompts, the attribute-enhanced context in soft prompt tuning may contain too much information about general information about a given class, hindering the alignment with a given image feature.
To better align with a given image feature and still exploit context attributes, we adapt the encoded query text feature $\tilde{\mathbf{t}}^a_i = g_T(\tilde{\mathbf{y}}^a_i)$ with an input-image-text-specific bias term $\boldsymbol{\beta}_i = g_M(\tilde{\mathbf{t}}^a_i, \tilde{\mathbf{f}})$.
Here, $g_M$ denotes a simple meta-network comprising multiple linear layers coupled with activation functions.
This process enhances the alignment between modalities by ensuring that the relevant class text features are better aligned with the corresponding image features. 
The class probability after adaptation is then represented as:
\begin{equation}
P(y=i|\mathbf{x}) = \frac{\exp\left(\cos(\tilde{\mathbf{f}},\tilde{\mathbf{t}}^a_i + \boldsymbol{\beta}_i) / \tau \right)}{\sum_{j=1}^{|C|} \exp\left(\cos(\tilde{\mathbf{f}},\tilde{\mathbf{t}}^a_j + \boldsymbol{\beta}_j) / \tau \right)}.
\end{equation}

\noindent
\textbf{Inference Only Vocabulary Ensemble.}
If we ask the LLM the same question multiple times, the answer may slightly vary each time.
Thus, attribute words generated from the LLM may vary for each sample output, leading to nontrivial performance variations.
During training, we just fix the vocabulary to a particular sample output.
But for inference on novel classes, we use three sample sets of $\mathbf{A}^{k}_{i}$, $k=\{1,2,3\}$. 
We then use the ensemble score from the average of each score from $\mathbf{A}^{k}_{i}$, as follows:
\begin{equation}
P_{novel}(y=i|\mathbf{x}) = \sum_{k=1}^K P(y=i|\mathbf{x},\mathbf{A}^{k}_{i})
\end{equation}

\section{Experiments}

\subsection{Settings} 
Following the seminal work~ \cite{CoCoOp}, the proposed CoAPT and other models are evaluated for three different few-shot classification tasks: base-to-novel generalization, cross-dataset transfer, and domain generalization. 


\noindent
\textbf{Base-to-Novel Generalization.} 
The base-to-novel generalization evaluation aims to assess the generalization of learning methods.
To do so, each dataset is first partitioned into two disjoint subsets: base and novel classes.
Then, examples of base classes are further divided into support sets for training and query sets for evaluation.
In particular, after training on few examples (support sets) of base classes, methods are assessed with respect to the performance on unseen examples (query sets) of seen base classes (base-to-base); the performance on unseen novel classes (base-to-novel); and the harmonic mean of the two.

\noindent
\textbf{Cross-Dataset Transfer.} 
To further evaluate the generalization of learning algorithms and models to unseen task distributions, they are first trained on a single dataset and then evaluated on diverse datasets with different classes, thus different task distributions.


\noindent
\textbf{Domain Generalization.} 
Domain generalization evaluation has a similar goal to cross-dataset transfer evaluation: evaluating the generalization to unseen distributions.
The difference is while cross-dataset transfer evaluation focuses on different task distribution (different classes), domain generalization evaluation looks at same task but different domains (e.g., styles).

\noindent
\textbf{Implementation Details.} Following the settings from~\cite{CoCoOp}, we employ the ViT-B/16 architecture of CLIP as a backbone architecture, and all experimental results are obtained by averaging results across three different random seeds. 
For each seed, our method generates three attribute word vocabulary from a large language model.
The results of baselines and other methods are  reproduced with their respective open-source code.

\begin{table*}[hbt!]
  \centering
  \footnotesize
  \centering\resizebox{\linewidth}{!}{
  \begin{tabularx}{\textwidth}{|l|XXX|XXc|XXc|XXc|}
    \hline
    \multirow{2}{*}{Method} & \multicolumn{3}{c|}{Avg. over 11 datasets} & \multicolumn{3}{c|}{ImageNet} & \multicolumn{3}{c|}{Caltech101} & \multicolumn{3}{c|}{OxfordPets} \\
    \cline{2-13}
    & Base & Novel & HM & Base & Novel & HM & Base & Novel & HM & Base & Novel & HM \\
    \hline
    CoOp~\cite{CoOp} & 82.69 & 63.22 & 71.66 & \textbf{76.47} & 67.88 & 71.92 & 98.00 & 89.81 & 93.73 & 93.67 & 95.29 & 94.47 \\
    +CoAPT & \textbf{82.70\textcolor{red}{\tiny{+0.01}}} & \textbf{72.79\textcolor{red}{\tiny{+9.57}}} & \textbf{77.43\textcolor{red}{\tiny{+5.77}}} & 76.32 & \textbf{68.96} & \textbf{72.45} & \textbf{98.28} & \textbf{92.83} & \textbf{95.48} & \textbf{94.74} & \textbf{97.58} & \textbf{96.13} \\
    \hline
    CoCoOp~\cite{CoCoOp} & 80.47 & 71.69 & 75.83 & \textbf{75.98} & \textbf{70.43} & \textbf{73.10} & 97.96 & \textbf{93.81} & \textbf{95.84} & \textbf{95.20} & \textbf{97.69} & \textbf{96.43} \\
    +CoAPT & \textbf{82.66\textcolor{red}{\tiny{+2.19}}} & \textbf{74.45\textcolor{red}{\tiny{+2.76}}} & \textbf{78.34\textcolor{red}{\tiny{+2.51}}} & 75.93 & 68.32 & 71.92 & \textbf{98.15} & 93.09 & 95.55 & 94.49 & 97.18 & 95.82 \\
    \hline
    KgCoOp~\cite{KG-CoOp} & 80.73 & 73.60 & 77.00 & 75.83 & 69.96 & 72.78 & 97.72 & \textbf{94.39} & \textbf{96.03} & \textbf{94.65} & \textbf{97.76} & \textbf{96.18} \\
    +CoAPT & \textbf{82.62\textcolor{red}{\tiny{+1.89}}} & \textbf{74.62\textcolor{red}{\tiny{+1.02}}} & \textbf{78.42\textcolor{red}{\tiny{+1.42}}} & \textbf{76.67} & \textbf{70.19} & \textbf{73.28} & \textbf{98.15} & 93.63 & 95.83 & 94.15 & \textbf{97.76} & 95.92 \\
    \hline
    MaPLe~\cite{MaPLe} & 82.28 & \textbf{75.14} & 78.55 & 76.66 & \textbf{70.54} & \textbf{73.47} & 97.74 & \textbf{94.36} & \textbf{96.02} & \textbf{95.43} & \textbf{97.76} & \textbf{96.58} \\
    +CoAPT & \textbf{83.77\textcolor{red}{\tiny{+1.49}}} & \text{75.13\textcolor{blue}{\tiny{\textbf{-0.01}}}} & \textbf{79.21\textcolor{red}{\tiny{+0.66}}} & \textbf{77.10} & 68.73 & 72.67 & \textbf{98.43} & 93.16 & 95.72 & 95.41 & 97.71 & 96.54 \\
    \hline
    PromptSRC~\cite{PromptSRC} & 84.26 & 76.10 & 79.97 & 77.60 & \textbf{70.73} & \textbf{74.01} & 98.10 & \textbf{94.03} & 96.02 & 95.33 & 97.30 & 96.30 \\
    +CoAPT & \textbf{84.74\textcolor{red}{\tiny{+0.48}}} & \textbf{77.07\textcolor{red}{\tiny{+0.97}}} & \textbf{80.72\textcolor{red}{\tiny{+0.75}}} & \textbf{77.87} & 69.88 & 73.66 & \textbf{98.43} & 93.81 & \textbf{96.06} & \textbf{95.59} & \textbf{97.71} & \textbf{96.63} \\
    \hline
    ProMetaR~\cite{ProMetaR} & \textbf{84.39} & 76.93 & 80.49 & \textbf{77.76} & \textbf{70.75} & \textbf{74.09} & 98.11 & 94.29 & 96.16 & 95.57 & 97.43 & 96.49 \\
    +CoAPT$\dagger$ & \text{84.00\textcolor{blue}{\tiny{\textbf{-0.39}}}} & \textbf{\underline{77.73}\textcolor{red}{\tiny{+0.80}}} & \textbf{80.74\textcolor{red}{\tiny{+0.25}}} & 77.54 & 70.63 & 73.92 & \textbf{98.41} & \textbf{94.61} & \textbf{96.47} & \textbf{95.62} & \textbf{97.69} & \textbf{96.64} \\
    \hline
    DePT~\cite{DePT} & \textbf{\underline{85.19}} & 76.17 & 80.43 & \textbf{78.20} & \textbf{70.27} & \textbf{74.02} & 98.57 & 94.10 & 96.28 & \textbf{95.43} & 97.33 & \textbf{96.37} \\
    +CoAPT & \text{84.92\textcolor{blue}{\tiny{\textbf{-0.27}}}} & \textbf{77.26\textcolor{red}{\tiny{+1.09}}} & \textbf{\underline{80.91}\textcolor{red}{\tiny{+0.48}}} & 76.88 & 67.20 & 71.72 & \textbf{98.77} & \textbf{94.76} & \textbf{96.72} & 94.67 & \textbf{97.52} & 96.07 \\
    \hline
    \multirow{2}{*}{Method} & \multicolumn{3}{c|}{StanfordCars} & \multicolumn{3}{c|}{Flowers102} & \multicolumn{3}{c|}{Food101} & \multicolumn{3}{c|}{FGVCAircraft} \\
    \cline{2-13}
    & Base & Novel & HM & Base & Novel & HM & Base & Novel & HM & Base & Novel & HM \\
    \hline
    CoOp~\cite{CoOp} & \textbf{78.12} & 60.40 & 68.13 & \textbf{97.60} & 59.67 & 74.06 & 88.33 & 82.26 & 85.19 & 40.44 & 22.30 & 28.75 \\
    +CoAPT & 76.44 & \textbf{70.83} & \textbf{73.53} & 97.37 & \textbf{74.04} & \textbf{84.09} & \textbf{89.89} & \textbf{90.73} & \textbf{90.31} & \textbf{40.86} & \textbf{38.47} & \textbf{39.63} \\
    \hline
    CoCoOp~\cite{CoCoOp} & 70.49 & \textbf{73.59} & 72.01 & 94.87 & 71.75 & 81.71 & \textbf{90.70} & \textbf{91.29} & \textbf{90.99} & 33.41 & 23.71 & 27.74 \\
    +CoAPT & \textbf{74.98} & 71.40 & \textbf{73.14} & \textbf{97.37} & \textbf{75.44} & \textbf{85.01} & 89.74 & 90.53 & 90.13 & \textbf{40.40} & \textbf{38.73} & \textbf{39.54} \\
    \hline
    KgCoOp~\cite{KG-CoOp} & 71.76 & \textbf{75.04} & 73.36 & 95.00 & 74.73 & 83.65 & \textbf{90.50} & \textbf{91.70} & \textbf{91.09} & 36.21 & 33.55 & 34.83 \\
    +CoAPT & \textbf{75.14} & 72.51 & \textbf{73.80} & \textbf{97.88} & \textbf{77.21} & \textbf{86.30} & 89.70 & 90.46 & 90.08 & \textbf{40.20} & \textbf{39.25} & \textbf{39.70} \\
    \hline
    MaPLe~\cite{MaPLe} & 72.94 & \textbf{74.00} & 73.47 & 95.92 & 72.46 & 82.56 & \textbf{90.71} & \textbf{92.05} & \textbf{91.38} & 37.44 & 35.61 & 36.50 \\
    +CoAPT & \textbf{77.59} & 72.41 & \textbf{74.90} & \textbf{97.40} & \textbf{77.68} & \textbf{86.43} & 89.91 & 90.71 & 90.31 & \textbf{41.16} & \textbf{38.31} & \textbf{39.67} \\
    \hline
    PromptSRC~\cite{PromptSRC} & 78.27 & \textbf{74.97} & \textbf{76.58} & \textbf{98.07} & 76.50 & 85.95 & 90.67 & 91.53 & 91.10 & 42.73 & 37.87 & 40.15 \\
    +CoAPT & \textbf{79.73} & 72.67 & 76.03 & 97.94 & \textbf{78.53} & \textbf{87.17} & \textbf{90.72} & \textbf{91.56} & \textbf{91.14} & \textbf{45.72} & \textbf{38.63} & \textbf{41.87} \\
    \hline
    ProMetaR~\cite{ProMetaR} & \textbf{78.32} & \textbf{75.18} & \textbf{76.72} & \textbf{98.13} & 77.66 & 86.70 & \textbf{90.80} & \textbf{91.89} & \textbf{91.34} & \textbf{42.02} & \textbf{38.63} & \textbf{40.25} \\
    +CoAPT$^\dagger$ & 76.80 & 74.64 & 75.70 & 97.69 & \textbf{80.14} & \textbf{88.05} & 90.76 & 91.75 & 91.25 & 40.94 & 38.31 & 39.56 \\
    \hline
    DePT~\cite{DePT} & 80.80 & \textbf{75.00} & \textbf{77.79} & 98.40 & 77.10 & 86.46 & \textbf{90.87} & 91.57 & \textbf{91.22} & 45.70 & 36.73 & 40.73 \\
    +CoAPT & \textbf{81.35} & 74.20 & 77.61 & \textbf{98.64} & \textbf{79.36} & \textbf{87.95} & 90.45 & \textbf{91.58} & 91.01 & \textbf{46.66} & \textbf{38.17} & \textbf{41.98} \\
    \hline
    \multirow{2}{*}{Method} & \multicolumn{3}{c|}{SUN397} & \multicolumn{3}{c|}{DTD} & \multicolumn{3}{c|}{EuroSAT} & \multicolumn{3}{c|}{UCF101} \\
    \cline{2-13}
    & Base & Novel & HM & Base & Novel & HM & Base & Novel & HM & Base & Novel & HM \\
    \hline
    CoOp~\cite{CoOp} & 80.60 & 65.89 & 72.51 & 79.44 & 41.18 & 54.24 & \textbf{92.19} & 54.74 & 68.69 & \textbf{84.69} & 56.05 & 67.46 \\
    +CoAPT & \textbf{81.03} & \textbf{74.56} & \textbf{77.66} & \textbf{81.17} & \textbf{54.87} & \textbf{65.48} & 89.35 & \textbf{59.44} & \textbf{71.30} & 84.23 & \textbf{78.40} & \textbf{81.20} \\
    \hline
    CoCoOp~\cite{CoCoOp} & 79.74 & \textbf{76.86} & \textbf{78.27} & 77.01 & 56.00 & 64.85 & 87.49 & 60.04 & 71.21 & 82.33 & 73.45 & 77.64 \\
    +CoAPT & \textbf{80.45} & 75.68 & 77.99 & \textbf{81.83} & \textbf{58.21} & \textbf{68.03} & \textbf{91.18} & \textbf{70.78} & \textbf{79.02} & \textbf{84.76} & \textbf{79.65} & \textbf{82.12} \\
    \hline
    KgCoOp~\cite{KG-CoOp} & 80.29 & 76.53 & 78.36 & 77.55 & \textbf{54.99} & \textbf{64.35} & 85.64 & 64.34 & 73.48 & 82.89 & 76.67 & 79.65 \\
    +CoAPT & \textbf{81.03} & \textbf{76.57} & \textbf{78.74} & \textbf{80.63} & 53.06 & 63.99 & \textbf{90.45} & \textbf{71.10} & \textbf{79.34} & \textbf{84.87} & \textbf{79.07} & \textbf{81.86} \\
    \hline
    MaPLe~\cite{MaPLe} & 80.82 & \textbf{78.70} & \textbf{79.75} & 80.36 & 59.18 & 68.16 & 94.07 & \textbf{73.23} & \textbf{82.35} & 83.00 & \textbf{78.66} & 80.77 \\
    +CoAPT & \textbf{81.55} & 76.27 & 78.82 & \textbf{82.41} & \textbf{61.55} & \textbf{70.35} & \textbf{95.33} & 71.50 & 81.65 & \textbf{85.18} & 78.38 & \textbf{81.61} \\
    \hline
    PromptSRC~\cite{PromptSRC} & \textbf{82.67} & \textbf{78.47} & \textbf{80.52} & 83.37 & 62.97 & 71.75 & 92.90 & 73.90 & 82.32 & \textbf{87.10} & 78.80 & 82.74 \\
    +CoAPT & 82.64 & 78.42 & 80.47 & \textbf{84.18} & \textbf{63.49} & \textbf{72.36} & \textbf{93.11} & \textbf{81.67} & \textbf{86.99} & 86.18 & \textbf{81.41} & \textbf{83.72} \\
    \hline
    ProMetaR~\cite{ProMetaR} & \textbf{82.70} & 79.02 & \textbf{80.82} & \textbf{83.02} & 64.05 & 72.31 & 94.94 & 77.44 & 85.30 & \textbf{86.97} & 79.84 & 83.25 \\
    +CoAPT$^\dagger$ & 82.36 & \textbf{79.22} & 80.76 & 82.79 & \textbf{64.73} & \textbf{72.65} & \textbf{95.41} & \textbf{81.24} & \textbf{87.75} & 85.68 & \textbf{82.06} & \textbf{83.83} \\
    \hline
    DePT~\cite{DePT} & \textbf{83.27} & \textbf{78.97} & \textbf{81.06} & \textbf{84.80} & 61.20 & 71.09 & 93.23 & 77.90 & 84.88 & \textbf{87.73} & 77.70 & 82.46 \\
    +CoAPT & 82.05 & 77.58 & 79.75 & 84.76 & \textbf{65.62} & \textbf{73.97} & \textbf{93.60} & \textbf{82.00} & \textbf{87.38} & 86.31 & \textbf{81.92} & \textbf{84.06} \\
    \hline
  \end{tabularx}}
  \caption{\textbf{Comparison of accuracy w/ or w/o CoAPT for base-to-novel generalization task.} We adapt the CoAPT method with 7 baseline models that do not fully utilize text encoder input. HM refers to harmonic mean. Bold letters indicate better performance between the baseline and CoAPT. Underlined text signifies the highest accuracy across methods. $\dagger$ denotes models without the meta-net $g_M$, only adding context attribute words during training and inference.}
  \label{base2novel}
\end{table*}

\noindent
\textbf{Datasets.} 
We follow the procedure from~\cite{CoCoOp} for constructing datasets.
For base-to-novel generalization and cross-dataset transfer tasks, we collect 11 different datasets, each constructed for different purposes and hence containing different data distributions. 
These include two generic-object datasets (ImageNet~\cite{imagenet} and Caltech101~\cite{caltech}); five fine-grained datasets (OxfordPets~\cite{oxfordpets}, StanfordCars~\cite{stanfordcars}, Flowers102~\cite{flowers102}, Food101~\cite{food101}, and FGVCAircraft~\cite{fgvcaircraft}); scene recognition dataset (SUN397~\cite{sun397}); action recognition dataset (UCF101~\cite{ucf101}); texture analysis dataset (DTD~\cite{dtd}); and satellite imagery dataset (EuroSAT~\cite{eurosat}).
For the domain generalization task, we take ImageNet as the primary source dataset, while taking its four variants as target datasets: ImageNetV2~\cite{imagenetv2}, ImageNet-Sketch~\cite{imagenet-sketch}, ImageNet-A~\cite{imagenet-a}, and ImageNet-R~\cite{imagenet-r}.
For all classes in all datasets, the support set comprises 16 shots, as sampled in~\cite{CoCoOp}.

\begin{table*}[htb!]
  \centering
  \scriptsize
  \begin{tabular}{lcccccccccccl}
    \toprule
    & Source & \multicolumn{10}{c}{Target} \\
    \cmidrule(lr){2-2} \cmidrule{3-13}
     & \rotatebox{45}{ImageNet} & \rotatebox{45}{Caltech101} & \rotatebox{45}{OxfordPets} & \rotatebox{45}{StanfordCars} & \rotatebox{45}{Flowers102} & \rotatebox{45}{Food101} & \rotatebox{45}{FGVCAircraft} & \rotatebox{45}{SUN397} & \rotatebox{45}{DTD} & \rotatebox{45}{EuroSAT} & \rotatebox{45}{UCF101} & \rotatebox{45}{Average} \\
    \midrule
    CoOp & \textbf{71.51} & \textbf{93.70} & 89.14 & \textbf{64.51} & 68.71 & \textbf{85.30} & 18.47 & 64.15 & 41.92 & \textbf{46.39} & 66.55 & 63.88 \\
    +CoAPT & \textbf{71.51} & 93.50 & \textbf{91.05} & 62.96 & \textbf{70.73} & 84.71 & \textbf{24.11} & \textbf{65.60} & \textbf{43.68} & 36.43 & \textbf{67.27} & \textbf{64.00\textcolor{red}{\tiny{+0.12}}} \\
    \midrule
    CoCoOp & \textbf{71.02} & \textbf{94.43} & 90.14 & \textbf{65.32} & \textbf{71.88} & \textbf{86.06} & 22.94 & \textbf{67.36} & 45.73 & \textbf{45.37} & 68.21 & \textbf{65.74} \\
    +CoAPT & 70.39 & 93.79 & \textbf{92.07} & 64.34 & 71.01 & 84.98 & \textbf{24.51} & 67.13 & \textbf{46.04} & 39.02 & \textbf{68.46} & 65.14\textcolor{blue}{\tiny{\textbf{-0.60}}} \\
    \midrule
    KgCoOp & 70.66 & 93.92 & 89.83 & 65.41 & 70.01 & \textbf{86.36} & 22.51 & 66.16 & 46.35 & \textbf{46.04} & 68.50 & 65.51 \\
    +CoAPT & \textbf{71.04} & \textbf{94.00} & \textbf{92.17} & \textbf{66.57} & \textbf{71.47} & 85.75 & \textbf{25.74} & \textbf{67.11} & \textbf{46.77} & 37.81 & \textbf{69.40} & \textbf{65.68\textcolor{red}{\tiny{+0.17}}} \\
    \midrule
    MaPLe & 70.72 & \textbf{93.53} & 90.49 & \textbf{65.57} & \textbf{72.23} & \textbf{86.20} & 24.74 & \textbf{67.01} & 46.49 & \textbf{48.06} & \textbf{68.69} & \textbf{66.30} \\
    +CoAPT & \textbf{71.87} & 93.09 & \textbf{92.11} & 64.28 & 70.02 & 84.86 & \textbf{25.22} & 66.11 & \textbf{46.73} & 47.37 & 67.82 & 65.76\textcolor{blue}{\tiny{\textbf{-0.54}}} \\
    \midrule
    PromptSRC & 71.27 & 93.60 & 90.25 & \textbf{65.70} & 70.25 & \textbf{86.15} & 23.90 & 67.10 & 46.87 & 45.50 & 68.75 & 65.81 \\
    +CoAPT & \textbf{71.35} & \textbf{94.23} & \textbf{92.18} & 65.29 & \textbf{70.82} & 86.06 & \textbf{25.63} & \textbf{67.22} & \textbf{47.87} & \textbf{53.38} & \textbf{68.91} & \textbf{\underline{67.16}\textcolor{red}{\tiny{+1.35}}}\\
    \midrule
    ProMetaR & \textbf{71.29} & 93.74 & 90.59 & \textbf{65.83} & \textbf{71.13} & \textbf{86.39} & 24.78 & \textbf{67.41} & 47.08 & 45.02 & \textbf{69.50} & 66.15 \\
    +CoAPT$^\dagger$ & 70.77 & \textbf{93.93} & \textbf{91.81} & 64.57 & 70.83 & 85.96 & \textbf{24.91} & 67.27 & \textbf{48.31} & \textbf{48.50} & 69.08 & \textbf{66.52\textcolor{red}{\tiny{+0.37}}} \\
    \midrule
    DePT & 71.60 & \textbf{93.80} & 90.13 & \textbf{66.00} & \textbf{70.93} & \textbf{86.27} & 24.30 & \textbf{67.23} & 46.60 & 45.83 & \textbf{69.10} & 66.02 \\
    +CoAPT & \textbf{71.73} & 93.67 & \textbf{91.36} & 64.62 & 70.09 & 85.79 & \textbf{24.68} & 67.07 & \textbf{48.58} & \textbf{51.45} & 69.06 & \textbf{66.64\textcolor{red}{\tiny{+0.62}}} \\
    \bottomrule
  \end{tabular}
  \caption{\textbf{Comparison of accuracy w/ or w/o CoAPT for cross-dataset transfer task.} Bold letters indicate better performance between the baseline and CoAPT. Underlined text signifies the highest accuracy across methods. $\dagger$ denotes model without the meta-net $g_M$.}
  \label{cross-tab}
\end{table*}

\subsection{Quantitative Evaluation}
To show the effectiveness, applicability, and flexibility of our CoAPT, we apply our CoAPT on top of several baselines: namely, CoOp~\cite{CoOp}, CoCoOp~\cite{CoCoOp}, KgCoOp~\cite{KG-CoOp}, MaPLe~\cite{MaPLe}, PromptSRC~\cite{PromptSRC}, ProMetaR~\cite{ProMetaR} and DePT~\cite{DePT}.
\textcolor{black}{
As ProMetaR already employs a meta-network to each text and image feature, we do not apply $g_m$ to ProMetaR to avoid redundancy.
}

\noindent
\textbf{Base-to-Novel Generalization} results of baselines and its CoAPT-enhanced version are presented in Table~\ref{base2novel} across 11 datasets.
We maintain all attribute words $\mathbf{A}_{i}$ and the meta-net architecture consistent across all baseline methods and datasets.
On average, CoAPT leads to improvements in the harmonic mean of the base and novel class accuracies for all baseline models. 
CoAPT leads to improvements for both base and novel class classification, except for MaPLe, ProMetaR and DePT.
The results demonstrate that combining enhanced context in hard prompts with soft prompt tuning can lead to better dataset adaptation and generalization.

\noindent
\textbf{Cross-Dataset Transfer} results are presented in Table~\ref{cross-tab}, evaluating the transferability of knowledge learned by models trained on ImageNet to various other datasets consisting of different classes. 
CoAPT generally improved the performance of baselines, especially achieving the largest performance improvement and thus the best results with PromptSRC as a baseline. 
It seems that the regularization process of PromptSRC, involving multiple template phrases as textual augmentation, can complement CoAPT more effectively than other baselines.
However, a slight decrease in performance is observed after applying CoAPT on CoCoOp and Maple.
Our conjecture is that this is due to their additional layers, such as meta-net architectures, which may introduce overfitting of the soft prompts.

\begin{table}[hbt!]
\centering
\scriptsize
\centering{
\begin{tabular}{lcccccl}
\toprule
 & \multicolumn{1}{c}{\textbf{Source}} & \multicolumn{4}{c}{\textbf{Target}} \\
\cmidrule(lr){2-2} \cmidrule(lr){3-7}
 & ImageNet & -V2 & -S & -A & -R & Avg. \\
\midrule
CoOp & \textbf{71.51} & \textbf{64.20} & 47.99 & 49.71 & 75.21 & 59.28 \\
+CoAPT & \textbf{71.51} & 64.18 & \textbf{48.25} & \textbf{50.53} & \textbf{75.28} & \textbf{59.56\textcolor{red}{\tiny{+0.28}}} \\
\midrule
CoCoOp & \textbf{71.02} & \textbf{64.07} & \textbf{48.75} & \textbf{50.63} & \textbf{76.18} & \textbf{59.91} \\
+CoAPT & 70.39 & 63.38 & 48.65 & 50.35 & 76.11 & 59.62\textcolor{blue}{\tiny{\textbf{-0.29}}} \\
\midrule
KgCoOp & \textbf{71.20} & 64.10 & 48.97 & 50.69 & 76.70 & 60.11 \\
+CoAPT & 71.04 & \textbf{64.27} & \textbf{49.14} & \textbf{51.08} & \textbf{76.87} & \textbf{60.34\textcolor{red}{\tiny{+0.23}}} \\
\midrule
MaPLe & 70.72 & 64.07 & \textbf{49.15} & \textbf{50.90} & \textbf{76.98} & \textbf{60.27} \\
+CoAPT & \textbf{71.87} & \textbf{64.72} & 49.01 & 47.78 & 76.43 & 59.49\textcolor{blue}{\tiny{\textbf{-0.78}}} \\
\midrule
PromptSRC & 71.27 & 64.35 & 49.55 & 50.90 & \textbf{77.80} & 60.65 \\
+CoAPT & \textbf{71.35} & \textbf{64.75} & \textbf{49.83} & \textbf{50.95} & 77.65 & \textbf{\underline{60.79}\textcolor{red}{\tiny{+0.14}}} \\
\midrule
ProMetaR & \textbf{71.29} & \textbf{64.39} & \textbf{49.55} & 51.25 & \textbf{77.89} & \textbf{60.77} \\
+CoAPT$\dagger$ & 70.77 & 64.32 & 49.52 & \textbf{51.41} & 77.59 & 60.71\textcolor{blue}{\tiny{\textbf{-0.06}}} \\
\midrule
DePT$^\ast$ & 71.58 & \textbf{64.32} & \textbf{48.82} & \textbf{49.74} & \textbf{76.76} & \textbf{59.91} \\
+CoAPT & \textbf{71.73} & 64.18 & 47.56 & 48.20 & 75.11 & 58.76\textcolor{blue}{\tiny{\textbf{-1.15}}} \\
\bottomrule
\end{tabular}}
\caption{{\textbf{Comparison of accuracy w/ or w/o CoAPT for domain generalization task.} } Bold letters indicate better performance between the baseline and CoAPT. Underlined text signifies the highest accuracy across methods. $\dagger$ denotes model without the meta-net $g_M$. $^\ast$ denotes our reproduced results.}
\label{domain-tab}
\end{table}

\noindent
\textbf{Domain Generalization} results are presented in Table~\ref{domain-tab}, evaluating the robustness of models to domain changes. 
CoAPT is shown to bring improvements upon CoOp, KgCoOp, and PromptSRC.
In particular, PromptSRC + CoAPT surpasses all other baselins and its CoAPT variants, showing the effectiveness of CoAPT.
However, for other baselines, CoAPT is observed to fail to bring improvements.
The accuracy drop may be due to the fact that attribute words tend to contain semantic information, which may not help to cope with different domains, such as styles of images and textures.



\begin{figure*}[hbt!]
  \centering
  \includegraphics[width=\textwidth]{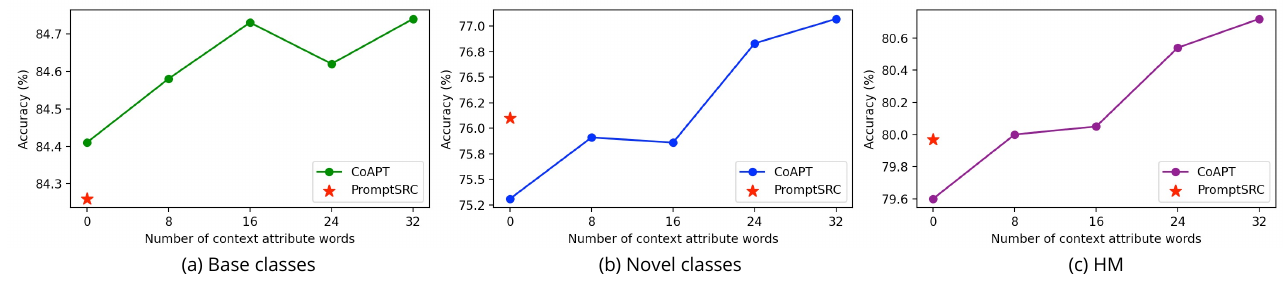}
  \caption{{\bf{Ablative evaluation of base-to-novel generalization on number of context attribute words with CoAPT integrated into PromptSRC~\cite{PromptSRC}}}. Plots are accuracy of base classes (a), novel classes (b), and harmonic mean (c), averaged over 11 datasets.}
  \label{att_num}
\end{figure*}

\subsection{Ablation Study}
In this section, we perform ablation studies to assess the effectiveness of each component of our proposed CoAPT.
We perform all ablation studies on PromptSRC as a main baseline in base-to-new generalization task.

\noindent
\textbf{Number of Attribute Words.}
We analyze the influence of the number of context attribute words on the performance, as shown in Figure~\ref{att_num}. 
No (zero) attribute word corresponds tosing the meta-net architecture without any attribute word.
With respect to all base-to-base accuracy, base-to-novel accuracy, and the harmonic mean of the two, a general trend of accuracy increase with the number of attribute words is observed.
In particular, we observe the highest improvement in the harmonic mean with 32 attribute words, which is the maximum number within the maximum context length.

\noindent
\textbf{CoAPT Subcomponents.}
To evaluate the contribution of each component of CoAPT to its overall performance, we sequentially add each component one at a time and assessed the impact on accuracy, as presented in Table~\ref{component-tab}. 
The positive effect of each component is evident, with the biggest portion coming from the combination of the context attributes with input specific query adaptation from the meta-net.
We also demonstrate the flexibility of CoAPT by integrating with the DePT~\cite{DePT} method, also added to a PromptSRC baseline, to achieve the maximum accuracy.

\begin{table}[t!]
    \centering
    \scriptsize
    \begin{tabular}{lccl}
        \toprule
        Method & Base Cls. & Novel Cls. & HM \\
        \midrule
        0: PromptSRC: reported in \cite{PromptSRC} & 84.26 & 76.10 & 79.97 \\
        1: PromptSRC: Reproduced & 84.24 & 75.81 & 79.80\\
        2: + Context Attribute Words & 84.56 & 75.95 & 80.03\textcolor{red}{\tiny{+0.23}} \\
        3: + Input Specific Query Adaptation & 84.74 & 76.99 & 80.68\textcolor{red}{\tiny{+0.65}} \\
        4: + Inference Vocabulary Ensemble & 84.74 & 77.07 & 80.72\textcolor{red}{\tiny{+0.04}}\\
        5: + DePT~\cite{DePT} & \textbf{84.92} & \textbf{77.26} & \textbf{80.91}\textcolor{red}{\tiny{+0.19}} \\
        \bottomrule
    \end{tabular}
    \caption{{\bf{Ablative evaluation of CoAPT components for base-to-novel generalization.}} Accuracy values are averaged over 11 datasets. HM refers to harmonic mean.}
    \label{component-tab}
\end{table}



\noindent
\textbf{Meta-net Configurations.} 
Table~\ref{method_table} presents the ablative evaluations on the meta-net configurations of CoAPT: namely, the output of the meta-net and where meta-net outputs are applied. 
In particular, we assess two options for the meta-net output: just a bias vector or affine transformation parameters (scale and bias vectors).
As for the location of application of the meta-net outputs, the outputs are either applied to text encoding $\tilde{\mathbf{t}}^a_i = g_T(\tilde{\mathbf{y}}^a_i)$ or soft prompts $\mathbf{P}^{T}$.

Our experiments indicate that adding bias to the text encoding $\tilde{\mathbf{t}}^a_i$ yields the highest performance. 
Interestingly, while the accuracy difference between bias and affine transformations when applied to $\mathbf{P}^{T}$ is small, the difference is much larger when applied to $\tilde{\mathbf{t}}^a_i$, by a margin of more than 2\% points. 
Perhaps the expansion of the controllable dimensions from the affine transformation may result in overfitting, rather than generalizability of soft prompts.

\begin{table}[t!]
    \centering
        \centering
        \small
        \begin{tabular}{lcccc}
            \toprule
            Output type & Applied to & Base Cls. & Novel Cls. & HM \\
            \midrule
            Affine & $\mathbf{P}^{T}$ & 84.53 & 76.97 & 80.57 \\
            Affine & $\tilde{\mathbf{t}}^a_i$ & 84.70 & 72.91 & 78.36 \\
            Bias & $\mathbf{P}^{T}$ & 84.32 & \textbf{77.25} & 80.63 \\
            Bias & $\tilde{\mathbf{t}}^a_i$ & \textbf{84.74} & 77.07 & \textbf{80.72} \\
            \bottomrule
        \end{tabular}
        \caption{{\bf{Ablative evaluation of meta-net configuration for base-to-novel generalization.}} $\tilde{\mathbf{t}}^a_i$ and $\mathbf{P}^{T}$ denote the text query encoding and soft prompt, respectively. Accuracy values are averaged over 11 datasets. HM refers to harmonic mean.}
        \label{method_table}
\end{table}

\begin{figure*}[t!]
  \centering
  \includegraphics[width=\textwidth]{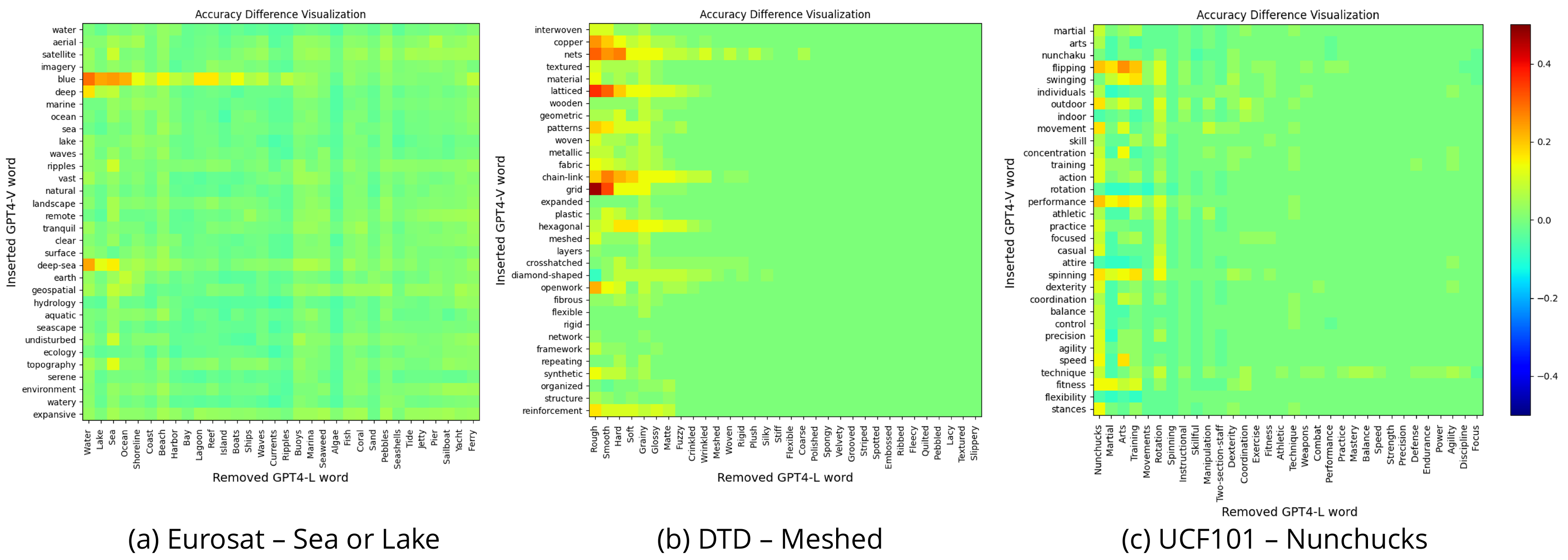}
  \caption{{\bf{Qualitative comparative analysis of attribute words from GPT4-Language and GPT4-Vision.}} Color denotes change in accuracy (scaled in range $[0,1]$) when replacing GPT4-Language generated word (x-axis) with that of GPT4-Vision (y-axis).}
  \label{word_change}
\end{figure*}

\noindent
\textbf{LLMs for Attribute Word Generation.} 
In Table~\ref{vocab_table}, we analyze how the performance is affected by a language model used to generate attribute words.
We choose LLaMa2-7B, LLaMa2-13B~\cite{touvron2023llama}, GPT4-Language, and GPT4-Vision~\cite{achiam2023gpt} as a possible candidate language model.
Serving as a pseudo oracle, when using GPT4-Vision as the attribute word generator, we provide images from the few-shot support set with customized instruction based on the class name for both base and novel classes.

We observe that the accuracy increases with the capacity of the large model, especially for novel classes. 
This indicates that the context attribute words indeed can positively enhance generalizability to novel classes, and that larger models tend to provide more relevant attribute words.

Another notable observation is the improvement from using GPT4-Vision, which is additionally provided with images of a support set to generate context attribute words.
While the use of the support set for novel classes violates the assumption of zero-shot classification, serving as pseudo oracle, it demonstrates that better attribute words can be generated with better input instructions. 
We believe that both these observations warrant further research in the attribute word generation process.


\begin{table}[t!]
        \centering
        \small
        \begin{tabular}{lccc}
            \toprule
            Vocabulary & Base Cls. & Novel Cls. & HM \\
            \midrule
            LLaMa2-7B & 84.62 & 75.29 & 79.68 \\
            LLaMa2-13B & 84.71 & 75.49 & 79.83 \\
            GPT4-Language & 84.74 & 77.07 & 80.72 \\
            \rowcolor{gray!15} 
            GPT4-Vision & \textbf{84.88} & \textbf{79.85} & \textbf{82.28} \\
            \bottomrule
        \end{tabular}
        \caption{{\bf{Ablative evaluation of attribute word generators for base-to-novel generalization.}} Accuracy values are averaged over 11 datasets. HM refers to harmonic mean.}
        \label{vocab_table}
\end{table}

\noindent
\textbf{Pairwise comparison of attribute words from GPT4-Language and GPT4-Vision.}
To investigate which specific attribute word may enhance or hinder classification performance, we perform qualitative analysis by replacing each attribute word generated using GPT4-Language with one that was generated using GPT4-Vision (a pseudo oracle), in Figure~\ref{word_change}.
Here, we exclude the meta-net for ease of comparison, as well as to exclusively examine the direct differences between wordsppwo. 
We guide the reader to focus on the red and orange colors in Figure~\ref{word_change}, which correspond to considerable improvements from the GPT-Vision generated attribute words.

For the Eurosat dataset, the classification accuracy of the class ``sea or lake'' considerably improves by inserting the words ``blue'', ``deep'', and ``deep-sea'' into the context. 
In the DTD dataset, the classification of the ``meshed'' class benefits from the words ``copper'', ``nets'', ``latticed'', ``chain-link'', and ``grid''. 
In the UCF101 dataset, the classification of the ``Nunchucks'' class is enhanced with the words ``flipping'', ``performance'', and ``spinning''. We can also see from the vertical patterns that the replacement of GPT4-Language word ``Nunchucks'', which is a duplicate of the class name, and words ``training'' and ``rotation'' with almost any GPT4-Vision word results in improvement.
While it is not clear-cut what the common characteristics between these words are, we can generally note that the GPT4-Vision words have more visual characteristics, in that they may be more readily visualized as an image.
For example, the words ``nets'', ``latticed'', and ``grid'' for the ``meshed'' class can be correlated to unambiguous visual structures, more so compared to GPT4-Language words ``rough'', ``smooth'', ``hard'' and ``soft'', to which many different visual descriptions can be associated.

\section{Conclusion}



We introduce CoAPT, a novel method that effectively integrates attribute words into the query context within soft prompt tuning in to enhance text queries in the CLIP embedding space. 
CoAPT is a plug-and-play method that can be seamlessly incorporated into various existing soft prompt tuning methods to improve the alignment between text and image embeddings. 
A meta-network is incorporated to compute feature biases that act as adaptation for the image-text combined queries.

Our experiments have shown that CoAPT can considerably improve accuracy across all few/zero-shot image classification tasks.
The results highlight the importance of using appropriate hard prompts within the input query context. 

The improvements from GPT4-Vision demonstrate that there are more effective attribute words. 
We believe our analysis on generating the optimal context attribute words may guide further research into the interactions between text and image latent spaces in pre-trained models.

{\small
\bibliographystyle{ieee_fullname}
\bibliography{egbib}
}

\clearpage
\onecolumn
\appendix
\section*{\centering \LARGE Appendix}
\section{Attribute Word Samples Generated by Various Foundation Models}
We leverage several foundation models~\cite{achiam2023gpt, touvron2023llama} to generate attribute words. 
Due to the inherent randomness and varying parameters of these models, the generated attribute words can differ. 
To maintain consistency in the attribute word results, we design a template instruction and apply it uniformly across all foundation models.
The general instruction is as follows: 

\leftskip=1cm
\rightskip=1cm
\vspace{3pt}
\noindent
``Please provide only \{Number of words\} descriptive English words for the \{Dataset name\} \{Class name\} dataset. Format your response as follows: $\backslash$nword$\backslash$nword$\backslash$nword$\backslash$nword''

\leftskip=0pt\rightskip=0pt
\noindent
We note that we have included explicit instructions to the model to generate English words only, as the model may generate words in other languages or even nonverbal outputs such as emoji.
We also include explicit instruction on the output format, for seamless data loading in downstream processes.

For the GPT-4 Vision model, instructions are given with the few-shot sample images, making the instructions slightly different, as follows:

\leftskip=1cm
\rightskip=1cm
\vspace{3pt}
\noindent
``Please take a look at this(these) image file(s) and provide only \{Number of words\} descriptive English words for the \{Dataset name\} \{Class name\} dataset. Format your response as follows:$\backslash$nword$\backslash$nword$\backslash$nword$\backslash$nword''
\vspace{3pt}

\leftskip=0pt\rightskip=0pt

Tables~\ref{sample_word_1} and \ref{sample_word_2} presents samples of attribute words for the base-to-novel dataset~\cite{imagenet, caltech, oxfordpets, stanfordcars, flowers102, food101, fgvcaircraft, sun397, dtd, eurosat, ucf101} generated by different foundation models. 
Full attribute words set and code available in : \url{https://github.com/LeeGun4488/CoAPT}

\begin{table}[t]
    \centering
    \begin{tabularx}{\textwidth}{|c|c|c|X|}
        \hline
        DB & Class & Model & Attribute word set \\
        \hline
        \multirow{8}{*}{\rotatebox{90}{ImageNet}} & \multirow{4}{*}{\rotatebox{90}{Goldfish}} & GPT4-V & Goldfish Aquarium Orange Fins Swimming Carassius Auratus Water \\
        \cline{3-4}
        & & GPT4-L & Goldfish Aquatic Fish Ornamental Pet Swimming Water Bowl \\
        \cline{3-4}
        & & LLaMa2-13B & Scales Fins Eyes Mouth Body Tail Colors Shapes \\
        \cline{3-4}
        & & LLaMa2-7B & Fish Gold Fishy Aquarium Swim School Tank Fin \\
        \cline{2-4}
         & \multirow{4}{*}{\rotatebox{90}{Castle}} & GPT4-V & Fortified Historic Medieval Towers Battlements Moat Stone Architecture \\
        \cline{3-4}
        & & GPT4-L & Medieval Fortress Palace Towers Gothic Stone Historic Majestic \\
        \cline{3-4}
        & & LLaMa2-13B & Majestic Medieval Towering Fortified Grandiose Historic Picturesque Sprawling \\
        \cline{3-4}
        & & LLaMa2-7B & Medieval Tower Moat Battlements Crenellations Drawbridge Curtain wall \\
        \hline
        \multirow{8}{*}{\rotatebox{90}{Caltech101}} & \multirow{4}{*}{\rotatebox{90}{Beaver}} & GPT4-V & Beaver Rodent Mammal Tail Fur Brown Water Aquatic \\
        \cline{3-4}
        & & GPT4-L & Animal Beaver Wildlife Nature Water Dam Canadian Aquatic \\
        \cline{3-4}
        & & LLaMa2-13B & Fur Tail Dam Pond Log Tree Branch Water \\
        \cline{3-4}
        & & LLaMa2-7B & Object Image Animal Beaver Water Scene Nature Aquatic \\
        \cline{2-4}
         & \multirow{4}{*}{\rotatebox{90}{Laptop}} & GPT4-V & Laptop Portable Computer Notebook Keyboard Screen Display Trackpad \\
        \cline{3-4}
        & & GPT4-L & Electronics Technology Computing Screen Hardware Laptop Gray Black \\
        \cline{3-4}
        & & LLaMa2-13B & Sleek Modern Slim Lightweight High-performance Powerful Efficient Fast \\
        \cline{3-4}
        & & LLaMa2-7B & Laptop Computer Device Gadget Technology Tool Machinery Equipment \\
        \hline
        \multirow{8}{*}{\rotatebox{90}{OxfordPets}} & \multirow{4}{*}{\rotatebox{90}{Beagle}} & GPT4-V & Beagle Dog Pet Animal Canine Tricolored Brown White \\
        \cline{3-4}
        & & GPT4-L & Adorable Alert Athletic Brown Curious Cute Domestic Energetic \\
        \cline{3-4}
        & & LLaMa2-13B & Fluffy Adorable Playful Curious Energetic Friendly Loyal Cuddly \\
        \cline{3-4}
        & & LLaMa2-7B & Adorable Playful Loyal Friendly Curious Energetic Affectionate Intelligent \\
        \cline{2-4}
         & \multirow{4}{*}{\rotatebox{90}{Bombay}} & GPT4-V & Bombay Cat Black Feline Glossy Yellow Eyes Domestic \\
        \cline{3-4}
        & & GPT4-L & Furry Shiny Playful Cute Sleek Affectionate Tuxedo Intelligent \\
        \cline{3-4}
        & & LLaMa2-13B & Furry Cute Playful Fluffy Adorable Friendly Curious Energetic \\
        \cline{3-4}
        & & LLaMa2-7B & Dog Cat Puppy Kitten Paw Fur Whiskers Tail \\
        \hline
        \multirow{8}{*}{\rotatebox{90}{StanfordCars}} & \multirow{4}{*}{\rotatebox{90}{\scriptsize{\shortstack{2012 Ferrari \\ 458 Italia Coupe}}}} & GPT4-V & Ferrari 458 Italia Coupe Red Yellow Luxury Sports \\
        \cline{3-4}
        & & GPT4-L & Ferrari 458 Italia Coupe 2012 Luxury Sports Speed \\
        \cline{3-4}
        & & LLaMa2-13B & Sleek Sporty Luxurious Red Shiny Powerful Aerodynamic High-performance \\
        \cline{3-4}
        & & LLaMa2-7B & Sleek Sporty Luxury Powerful Agile Aerodynamic Stylish Speedster \\
        \cline{2-4}
         & \multirow{4}{*}{\rotatebox{90}{\shortstack{2012 Audi \\S5 Coupe}}} & GPT4-V & Audi S5 Coupe 2012 Automobile Vehicle Luxury Performance \\
        \cline{3-4}
        & & GPT4-L & Luxurious Sleek Sporty Sophisticated Compact Powerful Stylish Modern \\
        \cline{3-4}
        & & LLaMa2-13B & Sleek Sporty Luxurious High-performance Powerful Agile Stylish Advanced \\
        \cline{3-4}
        & & LLaMa2-7B & Coupe Audi S5 Luxury Sports Sedan German Engine \\
        \hline
        \multirow{8}{*}{\rotatebox{90}{Flowers102}} & \multirow{4}{*}{\rotatebox{90}{\shortstack{Passion\\Flower}}} & GPT4-V & Passiflora Exotic Floral Radial Purple White Green Stigma \\
        \cline{3-4}
        & & GPT4-L & Passionflower Exotic Twining Climbing Tropical Subtropical Vibrant Multicolored \\
        \cline{3-4}
        & & LLaMa2-13B & Delicate Intricate Vibrant Exotic Colorful Intriguing Beautiful Detailed \\
        \cline{3-4}
        & & LLaMa2-7B & Petal Flower Vine Intricate Delicate Beauty Nature Bloom \\
        \cline{2-4}
         & \multirow{4}{*}{\rotatebox{90}{Rose}} & GPT4-V & Blooming Floral Petals Roses Gardening Botany Cultivation Fragrance \\
        \cline{3-4}
        & & GPT4-L & Rose Blooming Pink Red Floral Petals Garden Thorns \\
        \cline{3-4}
        & & LLaMa2-13B & Delicate Petal Velvety Red Full Bloom Soft Vibrant \\
        \cline{3-4}
        & & LLaMa2-7B & Beauty Bloom Colorful Delicate Elegant Fragrant Graceful Lovely \\
        \hline
        \multirow{8}{*}{\rotatebox{90}{Food101}} & \multirow{4}{*}{\rotatebox{90}{\shortstack{Chocolate\\Mousse}}} & GPT4-V & Chocolate Mousse Dessert Creamy Garnish Berries Whipped Cream \\
        \cline{3-4}
        & & GPT4-L & Tasty Chocolatey Sweet Creamy Rich Decadent Whipped Velvety \\
        \cline{3-4}
        & & LLaMa2-13B & Rich Creamy Decadent Smooth Silky Velvety Dark Bittersweet \\
        \cline{3-4}
        & & LLaMa2-7B & Rich Creamy Velvety Smooth Luxurious Decadent Dense Chocolatey \\
        \cline{2-4}
         & \multirow{4}{*}{\rotatebox{90}{Hamburger}} & GPT4-V & Beef Bun Sesame Lettuce Tomato Pickle Cheese Grilled \\
        \cline{3-4}
        & & GPT4-L & Bun Beef Cheese Lettuce Tomato Onion Pickle Mustard \\
        \cline{3-4}
        & & LLaMa2-13B & Juicy Savory Soft Meaty Tasty Cheesy Crispy Fresh \\
        \cline{3-4}
        & & LLaMa2-7B & Juicy Beef Bun Cheese Lettuce Tomato Onion Pickles \\
        \hline
    \end{tabularx}
    \caption{Sample attribute words per class in various datasets(Imagenet, Caltech101, OxfordPets, StanfordCars, Flowers102, Food101)}
    \label{sample_word_1}
\end{table}

\begin{table}[t]
    \centering
    \begin{tabularx}{\textwidth}{|c|c|c|X|}
        \hline
        DB & Class & Model & Attribute word set \\
        \hline
        \multirow{8}{*}{\rotatebox{90}{FGVCAircraft}} & \multirow{4}{*}{\rotatebox{90}{\shortstack{Global\\Express}}} & GPT4-V & Airplane Jet Wing Landing-gear Tail Fuselage Cockpit Windows \\
        \cline{3-4}
        & & GPT4-L & Global Express Aircraft Dataset FGVCAircraft Aerial Imagery Remote \\
        \cline{3-4}
        & & LLaMa2-13B & Aircraft Global Express Flight Travel Airline Passenger Journey \\
        \cline{3-4}
        & & LLaMa2-7B & Jet Aircraft Global Express Luxury Private Plane Flying \\
        \cline{2-4}
         & \multirow{4}{*}{\rotatebox{90}{Falcon 900}} & GPT4-V & Falcon 900 Aircraft Private Jet Trijet Landing Gear \\
        \cline{3-4}
        & & GPT4-L & Aircraft Aerospace Technology Jet Twin-engine Luxury Travel Transportation \\
        \cline{3-4}
        & & LLaMa2-13B & Luxurious Private Jet Plane Business Travel High-end Leather \\
        \cline{3-4}
        & & LLaMa2-7B & Jet Luxury Private Charter Flight Business Travel Aircraft \\
        \hline
        \multirow{8}{*}{\rotatebox{90}{SUN397}} & \multirow{4}{*}{\rotatebox{90}{Aquarium}} & GPT4-V & Aquarium Fish Water Coral Visitors Children Tunnel Marine \\
        \cline{3-4}
        & & GPT4-L & Aquatic Colorful Coral Fishes Tropical Marine Underwater Biodiversity \\
        \cline{3-4}
        & & LLaMa2-13B & Fish Tank Water Coral Seaweed Anemone Clownfish Goby \\
        \cline{3-4}
        & & LLaMa2-7B & Underwater Fish Tank Coral Reef Shark Sea Creatures \\
        \cline{2-4}
         & \multirow{4}{*}{\rotatebox{90}{Cemetery}} & GPT4-V & Tombstones Graves Cemetery Monuments Crosses Memorial Peaceful Trees \\
        \cline{3-4}
        & & GPT4-L & Grave Tombstone Cemetery Memorial Solemn Tranquil Peaceful Death \\
        \cline{3-4}
        & & LLaMa2-13B & Graveyard Tombstones Headstones Epitaphs Memorials Monuments Cenotaphs Obelisks \\
        \cline{3-4}
        & & LLaMa2-7B & Ancient Graves Headstones Tombstones Epitaphs Memorials Gravestones Markers \\
        \hline
        \multirow{8}{*}{\rotatebox{90}{DTD}} & \multirow{4}{*}{\rotatebox{90}{\shortstack{Dotted}}} & GPT4-V & Polka-dots Circles Patterns Textured Colorful Variegated Spotted Repetitive \\
        \cline{3-4}
        & & GPT4-L & Spotted Punctuated Speckled Dotted Pinpoint Stippled Freckled Mottled \\
        \cline{3-4}
        & & LLaMa2-13B & Bumpy Rough Dotted Speckled Spotted Splotchy Mottled Flecked \\
        \cline{3-4}
        & & LLaMa2-7B & Rough Smooth Bumpy Ridgy Prickly Silky Gritty Sticky \\
        \cline{2-4}
         & \multirow{4}{*}{\rotatebox{90}{Woven}} & GPT4-V & Interwoven Basketweave Textured Fabric Material Craftsmanship Handwoven Detailed \\
        \cline{3-4}
        & & GPT4-L & Rough Soft Patterned Threaded Grainy Coarse Smooth Ribbed \\
        \cline{3-4}
        & & LLaMa2-13B & Soft Rough Bumpy Smooth Fuzzy Velvety Shiny Dense \\
        \cline{3-4}
        & & LLaMa2-7B & Bumpy Rough Smooth Soft Silky Sandy Gritty Sticky \\
        \hline
        \multirow{8}{*}{\rotatebox{90}{EuroSAT}} & \multirow{4}{*}{\rotatebox{90}{\shortstack{Highway \\or Road}}} & GPT4-V & Aerial Highway Road Infrastructure Transportation Vehicles Asphalt Traffic \\
        \cline{3-4}
        & & GPT4-L & Asphalt Motorway Vehicles Transportation Urban Rural Traffic Suburban \\
        \cline{3-4}
        & & LLaMa2-13B & Road Highway Lane Path Pavement Asphalt Concrete Shoulder \\
        \cline{3-4}
        & & LLaMa2-7B & Road Highway Street Lane Junction Intersection Bridge Tunnel \\
        \cline{2-4}
         & \multirow{4}{*}{\rotatebox{90}{River}} & GPT4-V & Water River Aerial Satellite Imagery Curvature Meandering Flow \\
        \cline{3-4}
        & & GPT4-L & River Water Flow Stream Current Brook Rapids Outdoor \\
        \cline{3-4}
        & & LLaMa2-13B & River Water Stream Creek Watershed Channel Current Flow \\
        \cline{3-4}
        & & LLaMa2-7B & River Water Waters Flow Current Channel Shore Bank \\
        \hline
        \multirow{8}{*}{\rotatebox{90}{UCF101}} & \multirow{4}{*}{\rotatebox{90}{\shortstack{Playing\\Violin}}} & GPT4-V & Violin Performance Musicians Concert Classical Instrument Stringed Bow \\
        \cline{3-4}
        & & GPT4-L & Musical Violin Bow Strings Melodic Concentration Performance Instrument \\
        \cline{3-4}
        & & LLaMa2-13B & Skilled Gestures Movements Fingers Bowing Strings Posture Precision \\
        \cline{3-4}
        & & LLaMa2-7B & Violin Music Performance Notes Melody Rhythm Bowing Strings \\
        \cline{2-4}
         & \multirow{4}{*}{\rotatebox{90}{Surfing}} & GPT4-V & Surfing Waves Ocean Beach Surfers Surfboards Balance Riding \\
        \cline{3-4}
        & & GPT4-L & Surfing Ocean Waves Beach Surfboard Balance Splash Surfers \\
        \cline{3-4}
        & & LLaMa2-13B & Waves Ocean Beach Sun Sand Surf Board Ripples \\
        \cline{3-4}
        & & LLaMa2-7B & Wave Surf Ocean Beach Catch Ride Wiped Spray \\
        \hline
    \end{tabularx}
    \caption{Sample attribute words per class in various datasets(FGVCAircraft, SUN397, DTD, EuroSAT, UCF101)}
    \label{sample_word_2}
\end{table}

\end{document}